\documentclass[11pt, letterpaper]{article}

\usepackage{amsmath}
\usepackage{amssymb}
\usepackage{amsthm}
\usepackage{graphicx}
\usepackage{hyperref}
\usepackage{url}
\usepackage{algorithm}
\usepackage{algorithmic}
\usepackage{caption}
\usepackage{subcaption}
\usepackage{booktabs}
\usepackage{authblk}
\usepackage{appendix}
\usepackage[utf8]{inputenc}
\usepackage[T1]{fontenc}
\usepackage[english]{babel}
\usepackage{geometry}
\usepackage{xcolor}
\usepackage{fancyhdr}
\usepackage{microtype}
\usepackage{parskip}
\usepackage{tabularx}
\usepackage{enumitem}

\hypersetup{
    colorlinks=true,
    linkcolor=blue,
    filecolor=magenta,      
    urlcolor=cyan,
    citecolor=red,
}

\definecolor{headercolor}{RGB}{0, 50, 100}
\title{Radiology-GPT: A Large Language Model for Radiology}

\author[1]{Zhengliang Liu}
\author[2]{Aoxiao Zhong}
\author[1]{Yiwei Li}
\author[3]{Longtao Yang}
\author[3]{Chao Ju}
\author[1]{Zihao Wu}
\author[4]{Chong Ma}
\author[1]{Peng Shu}
\author[5]{Cheng Chen}
\author[5]{Sekeun Kim}
\author[1]{Haixing Dai}
\author[1]{Lin Zhao}
\author[6]{Lichao Sun}
\author[7]{Dajiang Zhu}
\author[3]{Jun Liu}
\author[8]{Wei Liu}
\author[9,10,11]{Dinggang Shen}
\author[5]{Xiang Li}
\author[5]{Quanzheng Li}
\author[1]{Tianming Liu}

\affil[1]{School of Computing, University of Georgia}
\affil[2]{Department of Electrical Engineering, Harvard University}
\affil[3]{Department of Radiology, Second Xiangya Hospital}
\affil[4]{School of Automation, Northwestern Polytechnical University}
\affil[5]{Department of Radiology, Massachusetts General Hospital and Harvard Medical School}
\affil[6]{Department of Computer Science and Engineering, Lehigh University}
\affil[7]{Department of Computer Science and Engineering, University of Texas at Arlington}
\affil[8]{Department of Radiation Oncology, Mayo Clinic}
\affil[9]{School of Biomedical Engineering, ShanghaiTech University}
\affil[10]{Shanghai United Imaging Intelligence Co., Ltd.}
\affil[11]{Shanghai Clinical Research and Trial Center}

\date{}

\begin{document}

\maketitle

\begin{abstract}
We introduce Radiology-GPT, a large language model for radiology. Using an instruction tuning approach on an extensive dataset of radiology domain knowledge, Radiology-GPT demonstrates superior performance compared to general language models such as StableLM, Dolly and LLaMA. It exhibits significant versatility in radiological diagnosis, research, and communication. This work serves as a catalyst for future developments in clinical NLP. The successful implementation of Radiology-GPT is indicative of the potential of localizing generative large language models, specifically tailored for distinctive medical specialties, while ensuring adherence to privacy standards such as HIPAA. The prospect of developing individualized, large-scale language models that cater to specific needs of various hospitals presents a promising direction. The fusion of conversational competence and domain-specific knowledge in these models is set to foster future development in healthcare AI. A demo of Radiology-GPT is available at \url{https://huggingface.co/spaces/allen-eric/radiology-gpt}.

\end{abstract}

\section{Introduction}
The recent rise of large language models (LLMs) such as ChatGPT \cite{liu2023summary} and GPT-4 \cite{openai2023gpt} has brought about a significant transformation in the field of natural language processing (NLP) \cite{zhao2023brain,liu2023summary}. These models demonstrate unprecedented abilities and versatility, a clear progression from preceding models such as BERT \cite{devlin2018bert}, and have engendered broad advancements in diverse domains \cite{liu2023summary,zhou2023comprehensive,dai2023chataug}. 

Among the fields that can significantly benefit from these advancements is radiology. Radiology, by its very nature, generates a vast amount of textual data, including radiology reports, clinical notes, annotations associated with medical imaging, and more \cite{ma2023impressiongpt}. Examples of such texts include radiographic findings, annotations for Computerized Tomography (CT) scans, and Magnetic Resonance Imaging (MRI) reports, all of which require sophisticated understanding and interpretation. 

Despite the transformative potential, the application of LLMs in the radiology domain has been limited  \cite{liu2023context,ma2023impressiongpt,wu2023exploring}. Large commercial models such as GPT-4 and PaLM-2 \cite{anil2023palm}, while powerful, are not readily applicable in clinical settings. HIPAA regulations, privacy concerns, and the necessity for IRB approvals pose substantial barriers \cite{liu2023deid}, primarily because these models necessitate the uploading of patient data to externally hosted platforms.

This situation underscores the pressing need for a localized foundational model, specifically designed for radiology, that can work effectively within the regulatory boundaries while capitalizing on the potential of LLMs. We address this gap through the development of Radiology-GPT, an LLM specifically designed for the radiology domain.

A crucial advantage of generative large language models such as Radiology-GPT, particularly in comparison to domain-specific BERT-based models such as RadBERT \cite{yan2022radbert} (for radiology) or ClinicalRadioBERT \cite{rezayi2022clinicalradiobert} (for radiation oncology), lies in their flexibility and dynamism. Unlike their BERT-based counterparts, generative LLMs are not rigidly dependent on a specific input structure, allowing for a more diverse range of inputs. Additionally, they are capable of generating diverse outputs, making them suitable for tasks that were previously considered impractical, such as reasoning \cite{wei2022emergent,zhong2023chatabl}. This further extends their versatility.

This inherent conversational capability of generative LLMs \cite{liu2023summary} positions them as invaluable aids to medical professionals, including radiologists. They can provide contextual insights and responses in a conversational manner, mimicking a human-like interaction, and hence enhancing the usability of these models in a clinical setting.

Furthermore, these models eliminate the need for complex fine-tuning processes and the associated labor-intensive manual annotation procedures. This attribute reduces both the development time and cost, making the adoption of such models more feasible and appealing in a clinical context. 

Radiology-GPT outperforms other instruction-tuned models not specially trained for radiology, such as Satbility AI's Stable LM \cite{stabilityStabilityLaunches} and Databrick's Dolly \cite{databricksFreeDolly}. The model is trained on the MIMIC-CXR dataset \cite{johnson2019mimic}, affording it a deep understanding of radiology-specific language and content. 

Key contributions of our work include:

\begin{itemize}
\item Development of a dedicated, localized large language model for the field of radiology, addressing the privacy and regulatory challenges in the clinical setting.
\item Radiology-GPT, leveraging the Alpaca instruction-tuning framework, demonstrates superior performance compared to general instruction-tuned models.
\item Training of the model on a large, richly annotated radiology dataset, ensuring its proficiency in handling complex radiological language and tasks.
\item Establishing a precedent for the development of localized foundational models in other medical specialties such as Radiation Oncology and Cardiology, encouraging further advancements in the application of LLMs in various healthcare domains.
\end{itemize} 

With these advancements, we believe Radiology-GPT holds immense potential to revolutionize the interpretation of radiology reports and other radiology-associated texts, making it a promising tool for clinicians and researchers alike.
\section{Related work}

\subsection{Large language models}
In recent years, the field of Natural Language Processing (NLP) has undergone a transformative shift with the advent of large language models (LLMs). These models, such as GPT-3 \cite{brown2020language}, GPT-4 \cite{openai2023gpt}, and PaLM \cite{chowdhery2022palm}, PaLM-2 \cite{anil2023palm}, offer enhanced language understanding capabilities, transforming the way language is processed, analyzed, and generated \cite{liu2023summary,zhao2023brain}. A notable shift in this evolution has been the transition from the \textbf{pre-training and fine-tuning} paradigm of BERT \cite{devlin2018bert}, GPT \cite{radford2018improving}, GPT-2 \cite{radford2019language} and their variants \cite{liu2019roberta,liao2023mask,zhou2023comprehensive}. In contrast, LLMs have introduced impressive few-shot and zero-shot learning capabilities achieved through in-context learning, offering a leap in both the performance and application scope of language models.

Alongside the growth of these powerful, commercially developed LLMs, the field has also seen the emergence of open-source alternatives. Models such as LLaMA \cite{touvron2023llama} and Bloom \cite{scao2022bloom} have extended the reach of LLMs, fostering democratization and inclusivity by offering the research community accessible and replicable models.

Building on open-source LLMs, the research community has begun to explore and develop instruction-tuned language models that can effectively interpret and execute complex instructions within the input text. Exemplifying this trend are models such as Alpaca \cite{stanfordStanfordCRFM}, StableLM \cite{stabilityStabilityLaunches}, and Dolly \cite{databricksFreeDolly}. Alpaca, fine-tuned from Meta’s LLaMA 7B model, has shown that even smaller models can achieve behaviors comparable to larger, closed-source models like OpenAI's text-davinci-003, while being more cost-effective and easily reproducible.

\subsection{Domain-specific language models}
Domain-Specific Language Models (DSLM) are models that are specifically trained or fine-tuned on text data pertaining to a particular domain or sector \cite{gu2021domain,liu2022survey}. By focusing on domain-relevant tasks and language patterns, these models seek to offer superior performance for tasks within their specialized domain.

For instance, AgriBERT \cite{rezayi2022agribert}, a domain-specific variant of BERT, is trained on an extensive corpus of text data related to agriculture and food sciences. It effectively captures specific jargon and language patterns typical of this sector, making it a valuable tool for tasks such as crop disease classification and food quality assessment.

In the field of education, SciEdBERT \cite{liu2023context} exemplifies the potential of DSLMs. SciEdBERT is pre-trained on student response data to middle school chemistry and physics questions. It is designed to understand and evaluate the language and reasoning of students in these domains, enabling accurate evaluation of student responses and providing insights into student understanding and misconceptions.

Similarly, in healthcare, ClinicalRadioBERT \cite{rezayi2022clinicalradiobert} has been developed for radiation oncology, training on clinical notes and radiotherapy literature to understand and generate radiation oncology-related text. These examples illustrate the diversity and potential of DSLMs in various sectors, enhancing performance on domain-specific tasks by leveraging tailored language patterns and specific knowledge.

While LLMs and DSLMs each have their strengths, there is a growing need for models that leverage the strengths of both, which is particularly pronounced in areas with specialized jargon and extensive bodies of text data, such as radiology.

\begin{figure}
    \centering
    \includegraphics[width=0.8\textwidth]{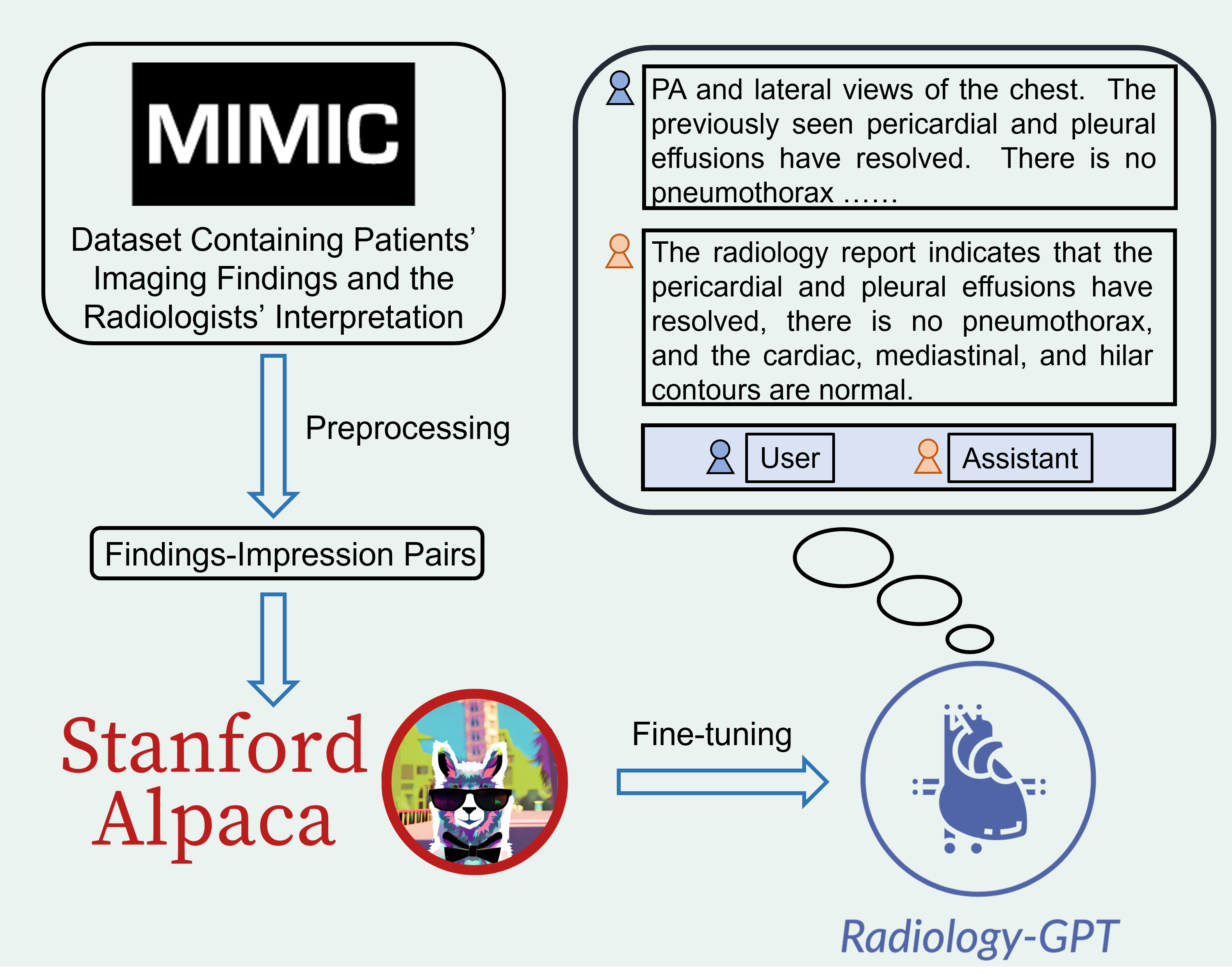}
    \caption{The overall framework of Radiology-GPT.}
    \label{fig:pipeline}
\end{figure}

\section{Methodology}
Our approach to building Radiology-GPT involves a two-step process: preprocessing of the dataset and fine-tuning the model with instruction following. The pipeline of our method can be seen in Figure \ref{fig:pipeline}.

\subsection{Dataset and preprocessing}
\label{sec:data}

We base our work on the publicly available MIMIC-CXR dataset \cite{johnson2019mimic}, a large, publicly available dataset of chest X-rays (CXRs). MIMIC-CXR contains de-identified medical data from over 60,000 patients who were admitted to the Beth Israel Deaconess Medical Center between 2001 and 2012.

We focus on the radiology reports available in this dataset, as they provide rich textual information about patients' imaging findings and the radiologists' interpretations. The reports typically contain two sections that correspond to each other: "Findings" and "Impression". The "Findings" section includes detailed observations from the radiology images, whereas the "Impression" section includes the summarized interpretations drawn from those observations.

To prepare the data for training, we preprocessed these reports to extract the "Findings" and "Impression" sections from each report and organized them into pairs. The preprocessing involved removing irrelevant sections, standardizing terminologies, and handling missing or incomplete sections. Specifically, we excluded ineligible reports through the following operations: (1) remove reports without finding or impression sections, (2) remove reports whose finding section contained less than 10 words, and (3) remove reports whose impression section contained less than 2 words.
We apply the official split published by \cite{johnson2019mimic} and finally obtain 122,014/957/1,606 reports for train/val/test set.

In addition, we also preprocessed the OpenI dataset \cite{demner2016preparing} based on the above exclusion operations to function as an independent external test dataset. It is crucial to validate our model's performance and generalizability across different data sources. Since the official split is not provided, we follow \cite{hu2021word} to randomly divide the dataset into train/val/test sets by 2400:292:576 (total: 3268 reports). The independent nature of the OpenI dataset allowed us to robustly assess our model's capabilities and understand its practical applicability. 
\begin{figure}
    \centering
    \includegraphics[width=0.9\textwidth]{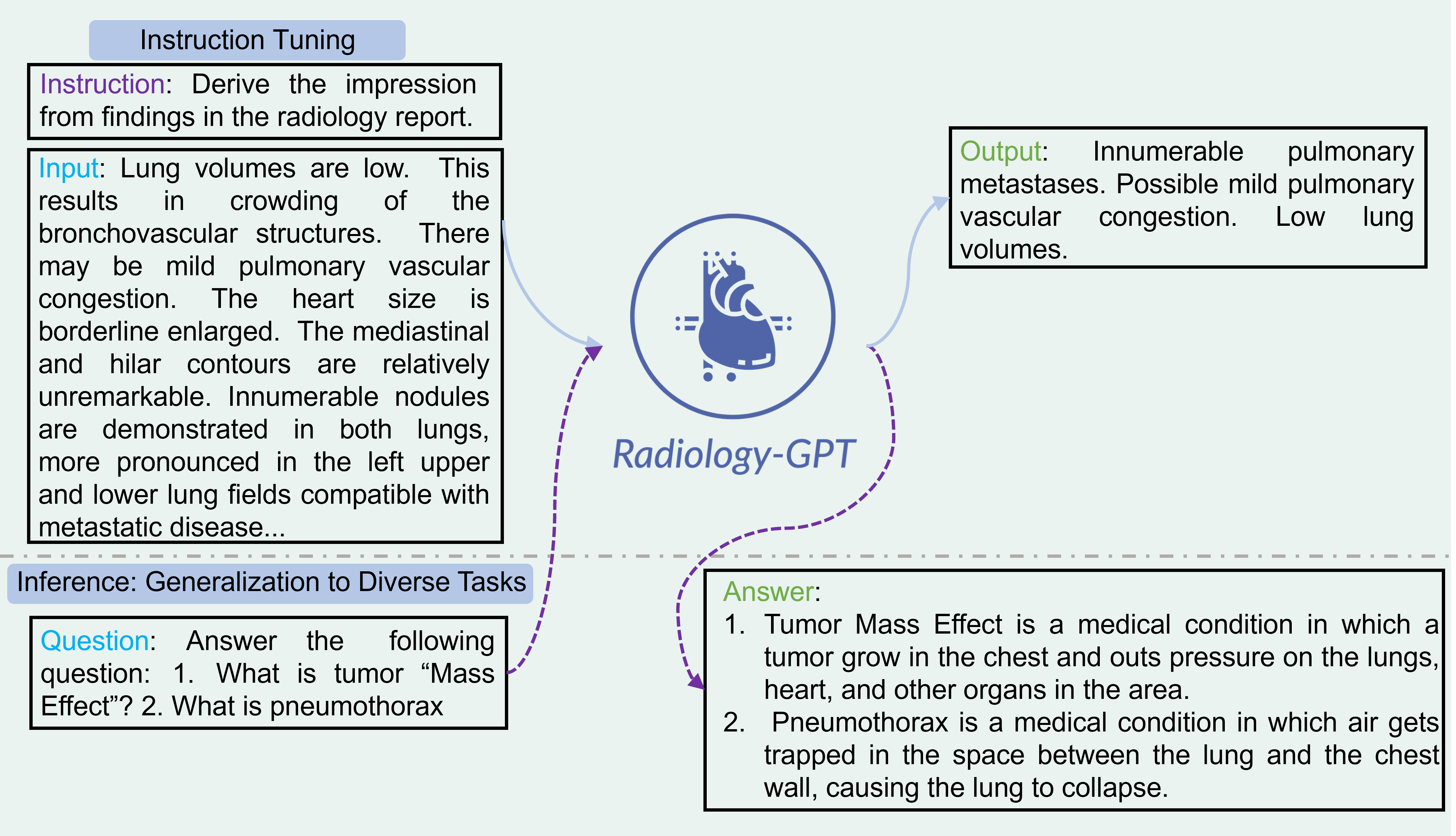}
    \caption{The instruction-tuning process of Radiology-GPT.}
    \label{fig:instruction_tuning}
\end{figure}
\begin{figure}
    \centering
    \includegraphics[width=\textwidth]{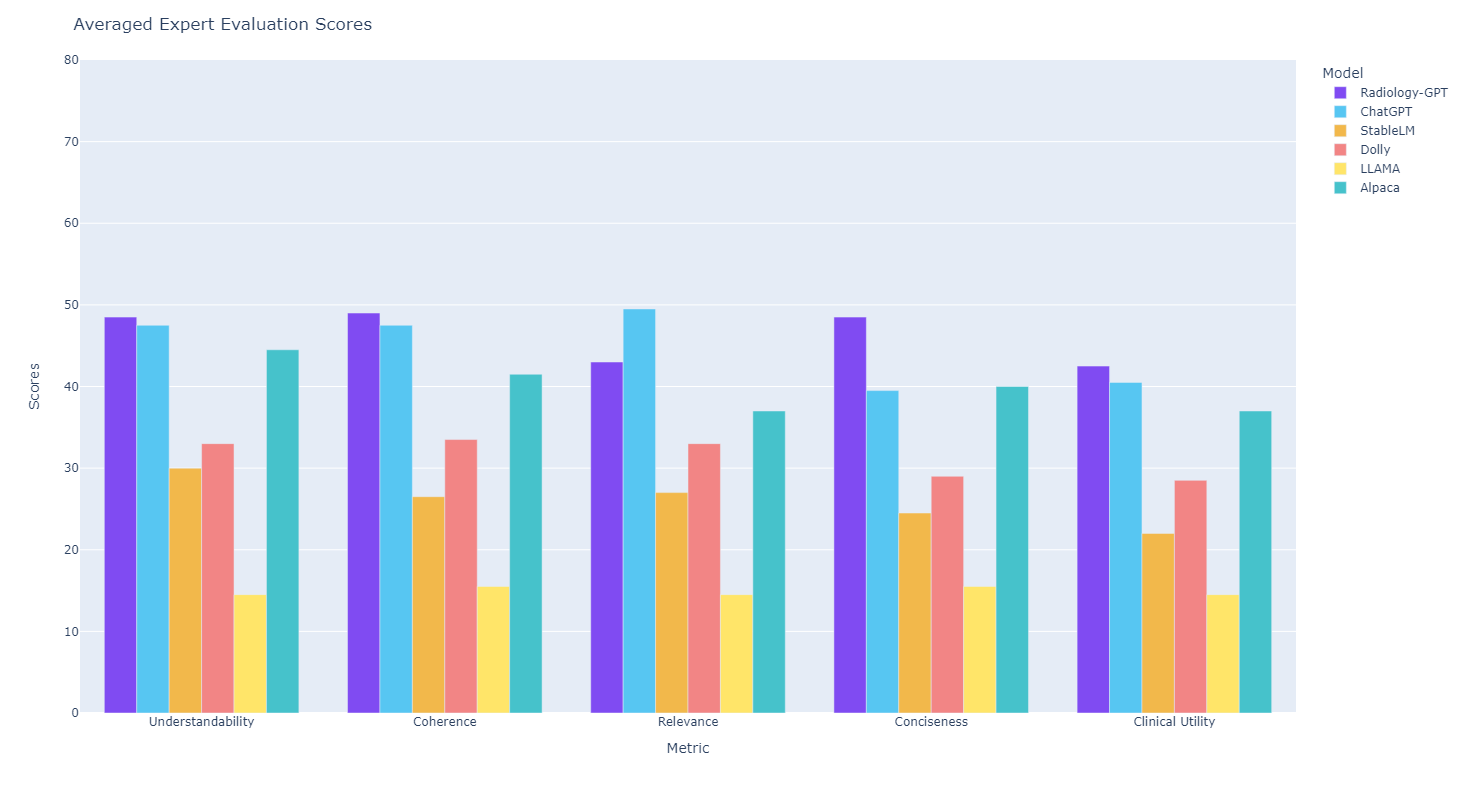}
    \caption{Comparisons of the LLMs based on Understandability, Coherence, Relevance, Conciseness, and Clinical Utility.}
    \label{fig:evaluation}
\end{figure}

\subsection{Experimental setting}
In this study, we trained the current version of Radiology-GPT based on Alpaca-7B. The training was conducted on a server with 4 Nvidia A100 80GB GPUs. We utilized LoRA \cite{hu2021lora} to facilitate training. More importantly, we followed the LoRA approach because the small size and portable nature of LoRA weights eases model sharing and deployment. 

The batch size was set to 128 and the learning rate was fixed at 3e-4. For LoRA, we set lora\_r, the rank of the low-rank factorization, to 8, and lora\_alpha, the scaling factor for the rank, to 16. This was accompanied by a dropout rate of 0.05. The target modules for LoRA were set to "q\_proj" and "v\_proj". These modules refer to the query and value matrices in the self-attention mechanism of the transformer architecture \cite{vaswani2017attention,hu2021lora}.

A very similar training process will be applied to larger versions of Radiology-GPT based on larger base models. 

\subsection{Instruction tuning}
\label{sec:fine-tuning}

The next step involves instruction tuning our LLM on this radiology dataset. Our methodology is based on the principle of instruction-tuning \cite{ouyang2022training,stanfordStanfordCRFM,wu2023interpretability}. The aim is to tune the model such that it can generate an "Impression" text given a "Findings" text as an instruction. The underlying language model learns the relationship between "Findings" and "Impression" from the dataset and hence, starts generating impressions in a similar manner when presented with new findings. Currently, we use Alpaca as the base model for this process. We will release versions of  Radiology-GPT in various sizes in the near future. The model also learns domain-specific knowledge relevant to radiology during this training process. 

To ensure our model learns effectively, we use a specific format for the instructions during training. We provide a short instruction to the "Findings" text: "Derive the impression from findings in the radiology report". The "Impression" text from the same report serves as the target output. This approach promotes learning by aligning the model with the task, thereby creating an instruction-following language model fine-tuned for radiology reports. Examples can be seen in Figure \ref{fig:instruction_tuning}.

Through this methodology, Radiology-GPT is designed to capture the specific language patterns, terminologies, and logical reasoning required to interpret radiology reports effectively, thereby making it an efficient and reliable tool for aiding radiologists.

It might be valuable to use diverse instruction pairs beyond "Findings ---> Impression" in radiology. Currently, "Findings ---> Impression" is the most natural and clinically meaningful task to conduct instruction-tuning. We are actively engaging with radiologists to construct a variety of clinically meaningful instruction tuning pairs to further enhance Radiology-GPT.

\section{Evaluation}
One of the challenging aspects of using large language models (LLMs) in the medical field, particularly in radiology, is the determination of their effectiveness and reliability. Given the consequences of errors, it's crucial to employ appropriate methods to evaluate and compare their outputs. To quantitatively evaluate the effectiveness of Radiology-GPT and other language models in radiology, we implemented a strategy to measure the understandability, coherence, relevance, conciseness, clinical utility of generated responses.

It should be noted that even trained radiologists have different writing styles \cite{wallis2011radiology,taira2001automatic}. There could be significant variations in how different radiologists interpret the same set of radiology findings \cite{elmore1994variability,gur2008laboratory,sonn2019prostate}. Therefore, we believe it is not appropriate to use string-matching \cite{alhendawi2013string}, BLEU \cite{papineni2002bleu}, ROUGE \cite{lin2004rouge}, or other n-gram based methods to evaluate the generated radiology impressions. Instead, we develop a set of metrics that are directly relevant to clinical radiology practices.

We describe the five metrics below:

\begin{itemize}
    \item Understandability: This metric assesses whether a human reader can understand the content generated by the LLM. Radiologists could rate the understandability of the generated impression section.
    \item Coherence: This metric assesses whether the LLM's output makes logical sense from beginning to end. For example, in a radiology report, it is necessary to evaluate whether the impression follows logically from the findings. 
    \item Relevance: This metric measures whether the impression generated by the LLM is relevant to the context and comprehensively covers core findings.
    \item Conciseness: This metric examines the succinctness of the LLM's output. The generated content should contain necessary and relevant information, without superfluous details or verbose explanations. 
    \item Clinical Utility: This is a key metric for any medical application. It evaluates whether the LLM's output is useful in a clinical context. Are the impressions generated actually useful for making diagnoses and treatment decisions?
\end{itemize}
\begin{figure}
    \centering
    \includegraphics[width=\textwidth]{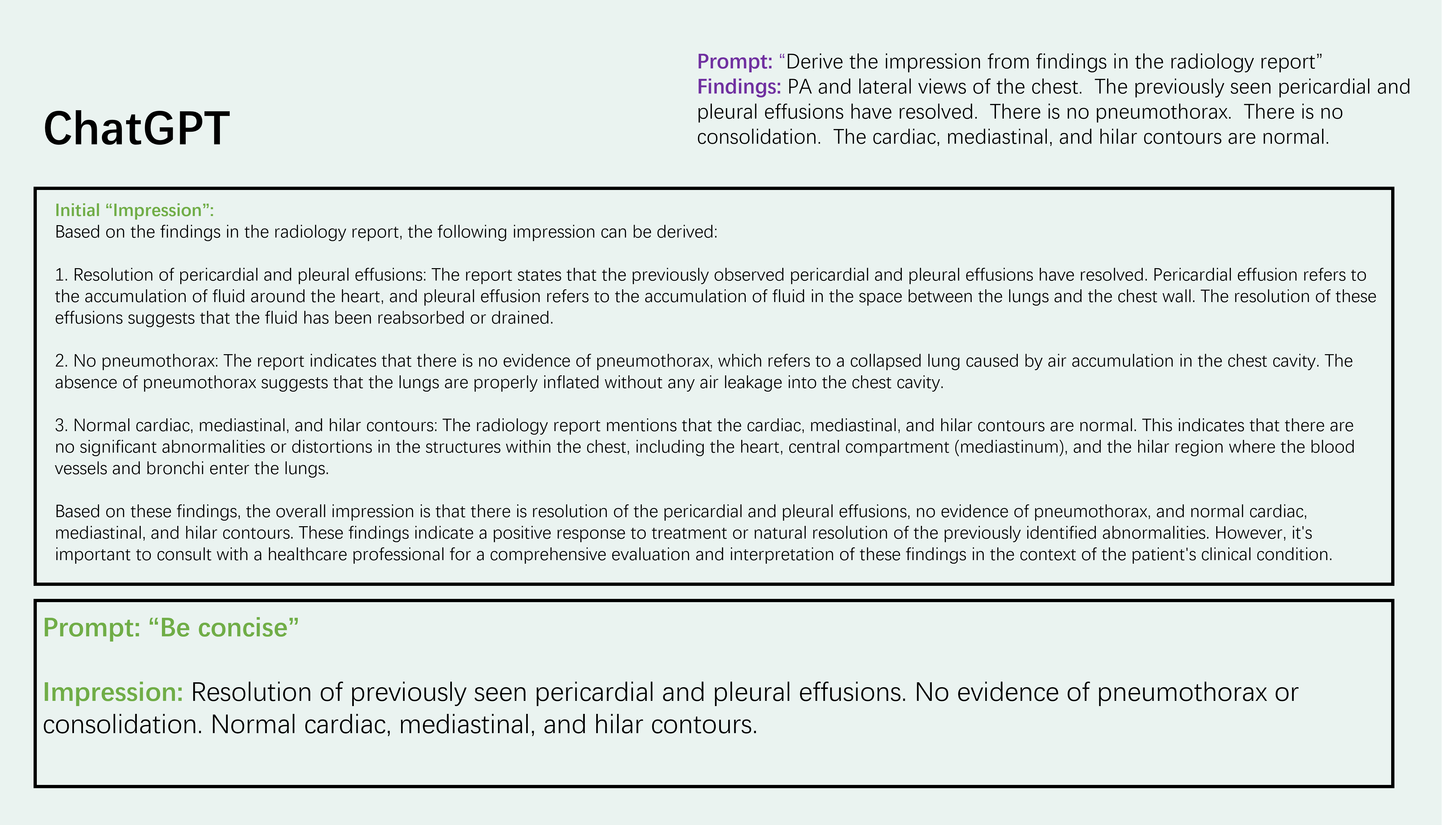}
    \caption{Results from ChatGPT.}
    \label{fig:model_comparison_chatgpt}
\end{figure}
\begin{figure}
    \centering
    \includegraphics[width=\textwidth]{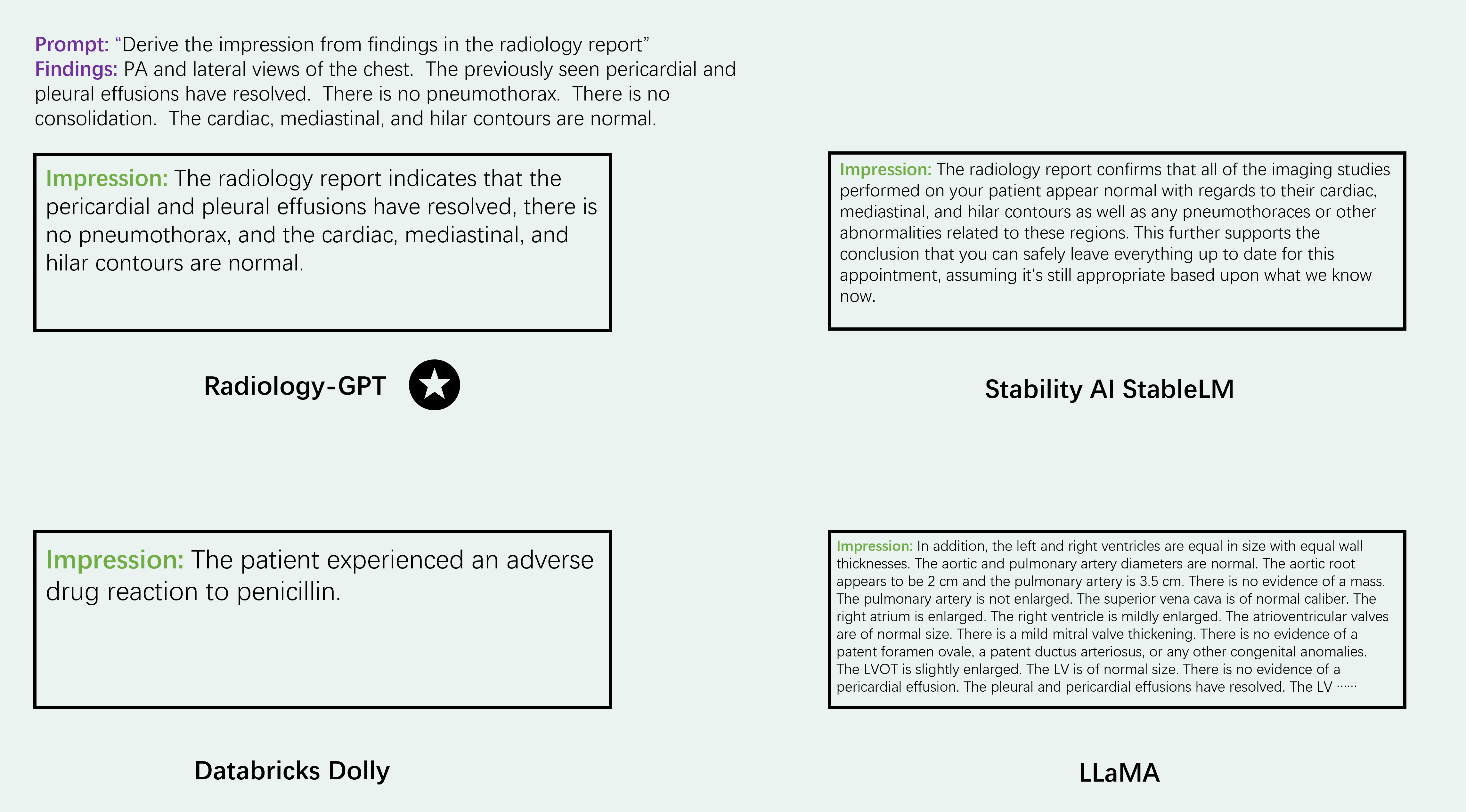}
    \caption{LLM comparison.}
    \label{fig:model_comparison_main}
\end{figure}

The experimental results yield insightful results regarding the performance of Radiology-GPT in comparison with other existing models. Figure \ref{fig:evaluation} shows the results of the expert evaluation of LLMs. A panel of two expert radiologists assessed the capacity of these LLMs in generating appropriate impressions based on given findings, considering five key parameters: understandability, coherence, relevance, conciseness, and clinical utility. Each metric was scored on a scale from 1 to 5. The radiologists independently assessed the quality of the LLMs' responses using a set of 10 radiology reports randomly selected from the test set of the MIMIC-CXR radiology reports and the test set of the OpenI dataset preprocessed by our in-house team. The assessments of the two radiologists were conducted independently, and the final scores for each metric were computed by averaging the scores from both radiologists, providing a balanced and comprehensive evaluation of each model's performance. 

The performance of Radiology-GPT was found to be comparable to that of ChatGPT in terms of understandability and slightly better in coherence. However, it lagged slightly behind in relevance, not due to a propensity to omit critical findings but rather, because it tended to produce shorter responses compared to ChatGPT. ChatGPT often generated lengthier responses that addressed nearly every point detailed in the findings (which sometimes can provide more context). Figure \ref{fig:model_comparison_chatgpt} shows the results of ChatGPT. This discrepancy could be attributed to the contrasting objectives of these models, with Radiology-GPT being designed to deliver succinct and focused outputs that capture the essential aspects of radiological impressions, a critical quality appreciated in the medical field. Consequently, this led to Radiology-GPT scoring higher in both conciseness and clinical utility. 

In contrast, the other tested models, including StableLM-7B, Dolly-12B, and LLaMA-7B, were all outperformed by both Radiology-GPT and ChatGPT. Figure \ref{fig:model_comparison_main} shows the results of these models. Despite Dolly-12B possessing a larger model size than Radiology-GPT 7B, it could not match the performance of our domain-tuned model. The lack of instruction tuning on domain-specific data and tasks within the field of radiology severely affected the performance of both StableLM-7B and Dolly-12B. 

LLaMA-7B, not having been instruction-tuned at all, struggled the most. It struggled to comprehend the given instructions and possesses insufficient domain-specific knowledge, leading to a markedly lower performance than other models. These findings underline the significant value that domain-specific tuning and instruction comprehension bring to the capabilities of LLMs in healthcare applications.

\begin{figure}
    \centering
    \includegraphics[width=\textwidth]{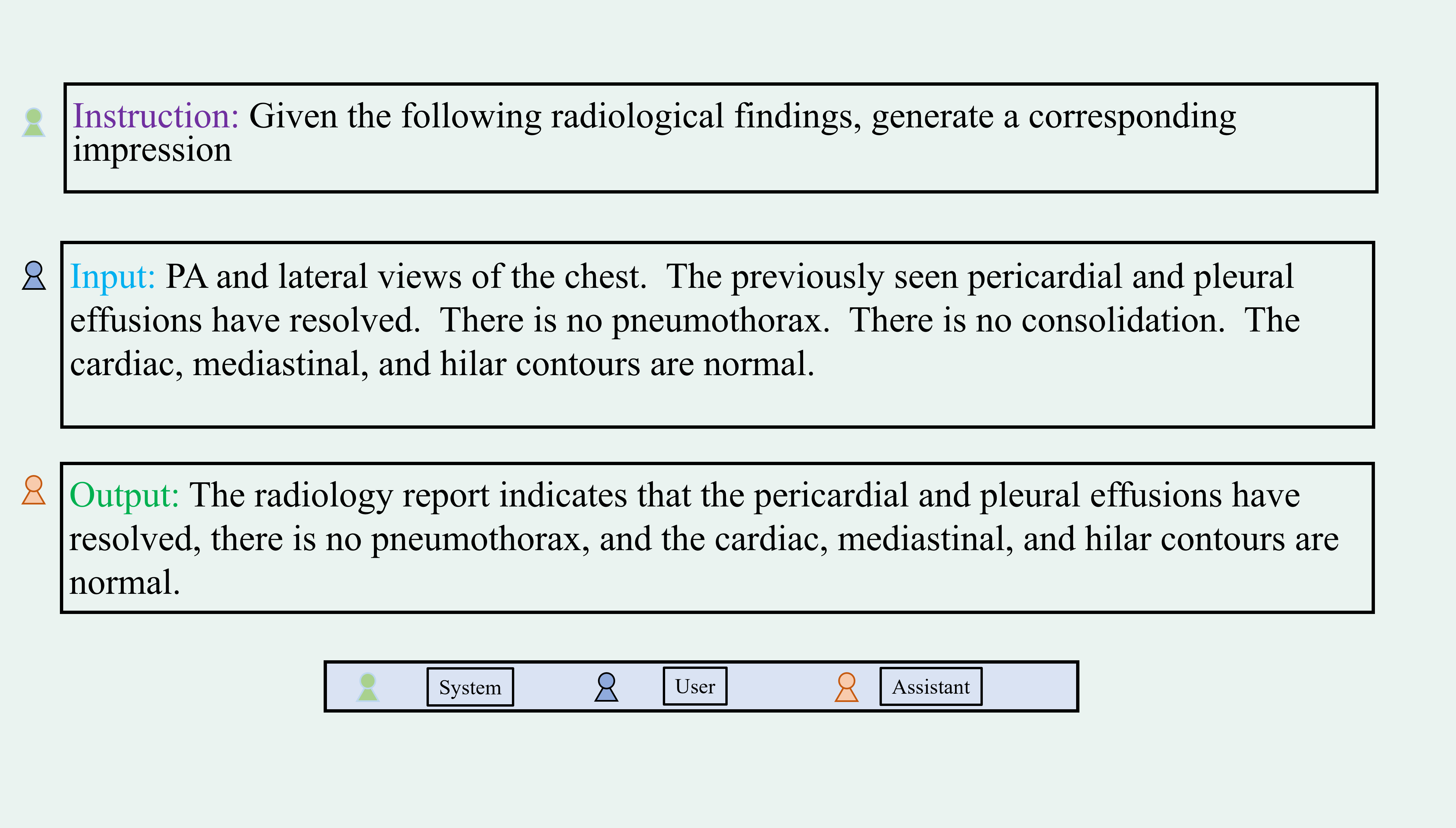}
    \caption{An example of deriving diagnostic impression from radiology findings.}
    \label{fig:example_2}
\end{figure}

\section{Discussion}
The exploration of large language models (LLMs), notably a locally deployable version specialized for radiology—hereafter referred to as Radiology-GPT—in the domains of medicine, and more pointedly in radiology, presents an intriguing trajectory with numerous potential future courses. In this section, we discuss future directions that are relevant to applying LLMs to radiology. 

\begin{figure}
    \centering
    \includegraphics[width=\textwidth]{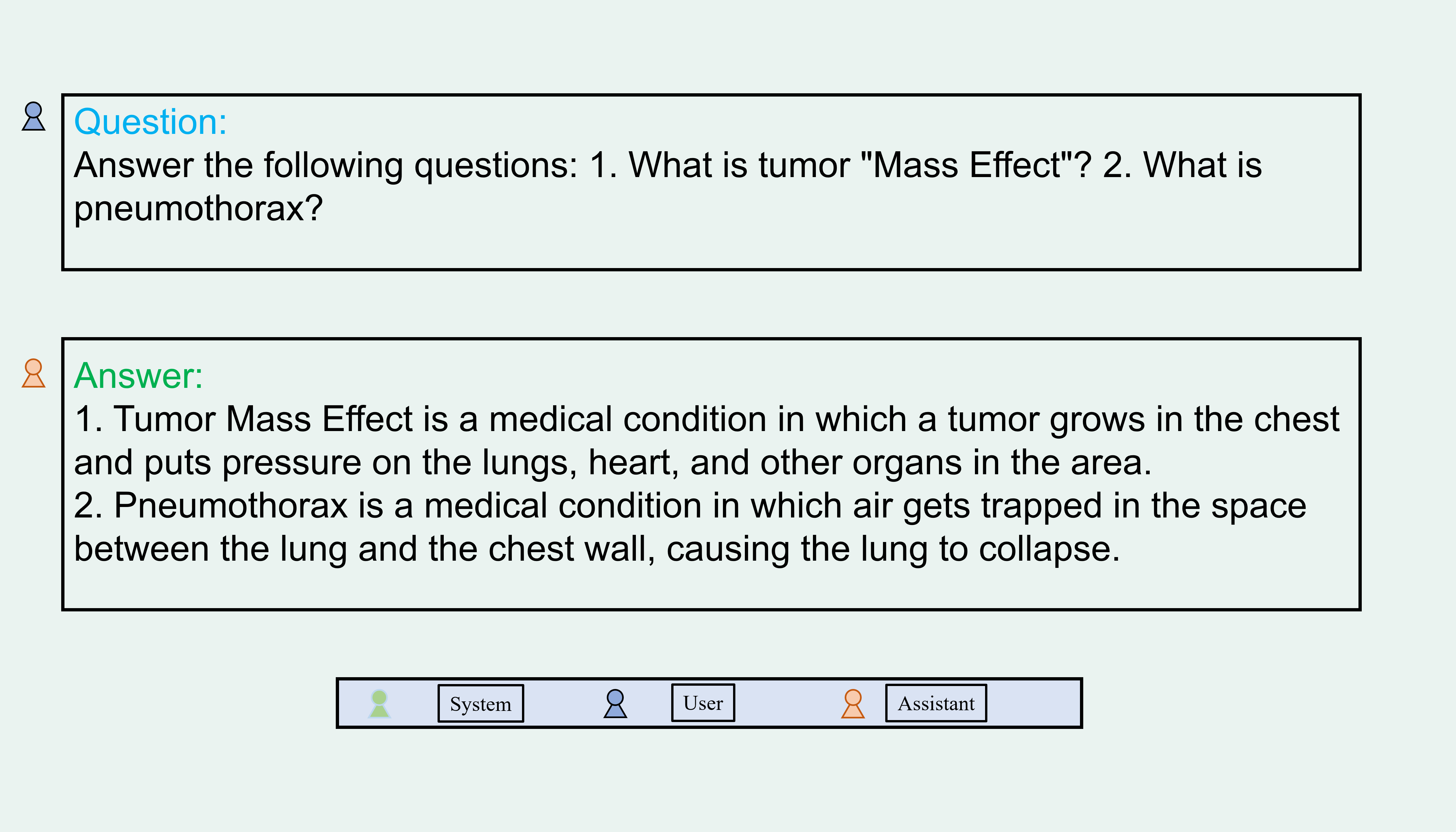}
    \caption{A conversational example.}
    \label{fig:example_1}
\end{figure}

\subsection{Clinical Decision Support}
Radiology-GPT can provide advice on various tasks, such as determining the best imaging modalities for specific clinical situations or assisting in the preparation of radiology reports. Figure \ref{fig:example_2} gives a perfect example. It's important to emphasize that the current generation of LLMs, including Radiology-GPT, ChatGPT, and GPT-4, are designed to augment professional judgment, not replace the roles of medical experts.

Future development should focus on amplifying LLMs' proficiency in deciphering intricate and multifaceted medical terminology, thus amplifying their relevance and reliability in clinical decision-making. The integration of Radiology-GPT with other AI modalities, such as computer vision models proficient in image interpretation, may well represent a significant progression.

\subsection{Augmentation of Patient Communication}

Another potential course is employing Radiology-GPT for the augmentation of patient communication. By transmuting complex medical vernacular and generating intelligible reports, this model could promote a higher level of understanding of medical information among patients. Figure \ref{fig:example_1} shows a conversational example. Nonetheless, its current capacity to produce precise and trustworthy patient communications without stringent supervision remains restricted.

Future endeavours should aim to refine Radiology-GPT to reduce inconsistencies and inaccuracies in its outputs and generate higher quality responses that are grounded in domain knowledge. This might be achievable through the fusion of enhanced domain-specific data and superior training strategies. Concurrently, it is crucial to engineer methods for effective validation and quality control of the generated communications prior to their delivery to patients.



\subsection{AGI Expert Panel}

In a vision of future healthcare scenarios, we envisage the deployment of localized domain-specific LLMs across different medical specialties that would coalesce into a virtual expert panel. Similar to interdisciplinary board panels of physicians who convene to discuss complex cases, such as those in oncology, these AI models could interactively aid in comprehensive decision-making.

In the context of oncology, for instance, a patient's case could be deliberated upon by a Radiology-GPT, a Pathology-GPT, a Oncology-GPT, and a Surgery-GPT, each providing insights from their respective domains. The Radiology-GPT might provide insights about the tumor size, its location, and spread based on the analysis of radiological scans. The Pathology-GPT could provide details on the cancer subtype and the cellular characteristics based on the biopsy report. The Oncology-GPT and Surgery-GPT might provide insights about potential treatment protocols and surgical options. All of these, combined together, could help in the creation of a comprehensive and personalized care plan.

While this concept presents significant potential to enhance the thoroughness of clinical decision-making, it also poses several challenges and questions. For instance, the coordination and interaction between these specialized LLMs would require a sophisticated communication framework. Moreover, there are potential risks associated with the aggregation of these insights, especially in cases of conflicting recommendations. Consequently, it would be critical to ensure human oversight to validate and reconcile the conclusions drawn by this AI panel.

Future research should focus on developing protocols for AI interaction, determining the most effective methods of integrating their insights, and establishing rigorous checks and balances to ensure the validity and reliability of the collective outputs. Equally significant, the development of such an integrated AI panel system should be guided by robust ethical frameworks to ensure that the autonomy and privacy of the patient are not compromised. 

Such an AI expert panel system represents a groundbreaking direction in the evolution of LLMs in healthcare, promising to harness the specialized knowledge of various medical fields to provide an unprecedented level of support in complex medical decision-making.

In summary, it is crucial to embrace the future of Radiology-GPT in medicine with both a degree of optimism and an attitude of circumspection. Active engagement with large language models, rigorous validation of their outputs, and ethical considerations, particularly relating to patient privacy and data usage, should guide their future evolution and deployment. This approach will ensure that the utilization of Radiology-GPT in healthcare will contribute positively to the progression of patient care, academic research, and the advancement of the medical profession as a whole.

\subsection{Privacy and ethics}
Privacy and ethics are significant considerations when integrating AI technologies in health care.  Radiology-GPT, given its nature as a locally trained and locally deployed model, upholds patient privacy. This is an area where local models show their superiority over commercially developed LLMs such as ChatGPT or PaLM-2 \cite{anil2023palm}. By operating within the confines of the hosting hospital's servers, Radiology-GPT mitigates the risk of patient Protected Health Information (PHI) leaks, making the approach compliant with the Health Insurance Portability and Accountability Act (HIPAA) regulations \cite{liu2023deid}.

However, it is crucial to acknowledge the potential ethical risks that come with the deployment of such models. One of these risks is the possibility of dispensing inaccurate or ungrounded medical advice \cite{sallam2023utility}. The application of LLMs in a medical setting could inadvertently misinform or misdirect care, especially in the absence of appropriate regulation and control mechanisms. Thus, it is essential to ensure rigorous oversight and regular auditing of these models' performance.

Additionally, bias in training data is a persistent concern in AI, with repercussions potentially amplified in the healthcare context \cite{kelly2019key}. If the data used to train Radiology-GPT reflects systematic bias in patient treatment or diagnosis, the model could propagate and even exacerbate these biases. Consequently, fairness, accountability, and transparency in model training are paramount to avoid such pitfalls.

It is also necessary to consider the risk of exploitation of these models to generate counterfeit content or misinformation \cite{liao2023differentiate}. The impressive language generation capabilities of LLMs can be manipulated to create misleading or entirely false health information, with potential consequences for public health. 

\subsection{Expanding Instruction Tuning Data in Radiology }
In our current work, we utilized "Findings ---> Impression" pairs as our primary instruction tuning data. While this forms a crucial part of clinical radiology, the potential utility of Radiology-GPT can be greatly enhanced by constructing and incorporating more diverse and clinically meaningful instruction pairs.

In our work, "Findings ---> Impression" serves as a natural starting point, mimicking the real-life task of radiologists interpreting findings to provide an impression. However, the field of radiology is rich with a multitude of other tasks and workflows that could potentially be translated into instruction pairs for Radiology-GPT. 

For instance, instructions could be formulated to generate more patient-friendly explanations of radiology reports or to write up recommendations for further diagnostic tests or treatment based on specific findings. Another potential use case could be to instruct Radiology-GPT to generate a list of differential diagnoses based on presented findings, or to summarize the latest research on specific imaging findings or diseases.

Moreover, as radiology practices involve a considerable amount of protocol decision-making \cite{glazer2018ct,einstein2014patient}, Radiology-GPT could also be tuned to suggest optimal imaging protocols based on a given clinical scenario. 

Expanding the range of instruction pairs can potentially enhance Radiology-GPT's versatility, making it a more holistic tool that can address different aspects of the radiology workflow. 

To achieve this, active engagement with practicing radiologists is crucial. By collaborating with experts in the field, we can identify a broad range of real-world tasks that Radiology-GPT can be trained to perform, ultimately ensuring that the model's capabilities are aligned with the needs of clinical practice.

\section{Conclusion}

In conclusion, we have developed Radiology-GPT, a domain-specific large language model that addresses the critical need for a locally deployable AI solution within the field of radiology. By leveraging the strength of large language models and tailoring it to radiological contexts, Radiology-GPT offers a promising leap forward, displaying superior performance over existing baselines in our evaluations. The evaluation metrics we proposed, which encapsulate a combination of qualitative and quantitative measures, offer a robust framework for assessing the effectiveness of this and similar models within healthcare settings. 

Moreover, Radiology-GPT opens up exciting avenues for future applications. It provides a foundation for the incorporation of multimodal data, including radiological images, further enhancing its potential contributions to the field. Its localized nature also paves the way for wider applications in other medical specialties, stimulating advancements in healthcare AI that respect privacy regulations. Our study stands as a testament to the potential of specialized AI in medicine, offering both immediate benefits and laying the groundwork for future innovation.

\bibliography{LLM_refs}
\bibliographystyle{unsrt}

\end{document}